\documentclass{svproc}
\usepackage{url}
\usepackage{amssymb}
\usepackage{amsfonts}
\usepackage{amsmath}
\usepackage{algorithm,algorithmic,caption}
\usepackage{subcaption}
\usepackage{graphicx}
\usepackage{color}
\usepackage{enumerate}
\usepackage{float,framed,url}

\begin{document}
\mainmatter              
\title{Investigating the Impact of Data Volume and Domain Similarity on Transfer Learning Applications}
\titlerunning{Impact of Data Volume and Domain Similarity in Transfer Learning}  
%
\author{Michael Bernico \and Yuntao Li \and Dingchao Zhang}
\authorrunning{M. Bernico et al.} 
%
\tocauthor{Ivar Ekeland, Roger Temam, Jeffrey Dean, David Grove,
Craig Chambers, Kim B. Bruce, and Elisa Bertino}
\institute{State Farm Insurance, Bloomington IL 61710, USA,\\
\email{michael.bernico.qepz@statefarm.com}}
\maketitle              

\begin{abstract}
Transfer learning allows practitioners to recognize and apply knowledge learned in previous tasks (source task) to new tasks or new domains (target task), which share some commonality. The two important factors impacting the performance of transfer learning models are: (a) the size of the target dataset, and (b) the similarity in distribution between source and target domains. Thus far, there has been little investigation into just how important these factors are. In this paper, we investigate the impact of target dataset size and source/target domain similarity on model performance through a series of experiments. We find that more data is always beneficial, and model performance improves linearly with the log of data size, until we are out of data. As source/target domains differ, more data is required and fine tuning will render better performance than feature extraction. When source/target domains are similar and data size is small, fine tuning and feature extraction renders equivalent performance. Our hope is that by beginning this quantitative investigation on the effect of data volume and domain similarity in transfer learning we might inspire others to explore the significance of data in developing more accurate statistical models.
\keywords{Computer Vision, Deep Learning, Transfer Learning, Business Application, Domain Similarity, Data Volume Effect}
\end{abstract}

\section{Introduction}
For many applications of computer vision, data is in short supply.  While Convolutional Neural Networks (ConvNets) trained on large datasets such as ImageNet \cite{ref1} with the benefit of massive computational power have revolutionized computer vision, many business applications of computer vision has limited relevant data.  Many researchers and machine learning engineers have had great success in addressing this problem by utilizing transfer learning \cite{ref2}.  

With transfer learning the knowledge from a network trained on a large dataset such as ImageNet (source domain) are transferred to another problem domain (target domain).  A variety of techniques can be applied to accomplish this, however, most commonly the final network layers of the original network are replaced with layers more suitable for the target domain, and the network is then trained on data related to that target domain.  This network benefits from the learnings of the original network, allowing the network to perform its task using significantly less training data.

In the rest of this paper, we compare and analyze data collected via training a variety of neural networks, using different training methods and open source datasets, to investigate and present the relationship between data volumes and source, target domain similarity in relates to model performance.

\subsection{Related Work}
This study is motivated nearly entirely by the work done by \cite{ref3}, in ``Revisiting Unreasonable Effectiveness of Data in Deep Learning Era.'' In this paper, the authors notes that while model size and computational power have increased over the previous 5 years, ImageNet is still being used to train those models. The paper goes on to examine the impact of increasing the volume of training data from ImageNet's 10 million images to 300 million images in the Google JFT-300M image dataset. In doing so, the authors observed that model performance increases logarithmically based on the volume of training data.  These results sparked our curiosity.  We wanted to understand the impact that target domain data volume and source/target domain similarity had on model performance.

\section{Experiments}
To understand the impact of data volumes, and relationships between target and source data domains in transfer learning problems, various pre-trained neural network architectures are selected and further tuned with new training data of different domains and sizes over a series of experiments. Model performances are subsequently compared and analyzed to obtain more insights.
\subsection{Datasets}
Two datasets are primarily used throughout this study, namely the MiniPlace dataset and IMDB-WIKI dataset, although internally enhanced and preprocessed. The Kaggle's Dogs vs. Cats (Dogs/Cats) data are also used as an exploratory dataset.  

The Dogs/Cats dataset is from Kaggle's Dogs vs. Cats competition, originally provided by Microsoft Research\cite{ref4}. The dataset consists of 25,000 training images of cats and dogs, creating a binary classification problem. A separate unlabeled test set was provided but not used.  

The MiniPlace data set originates from the MIT Places2 data set\cite{ref7}, a scene-centric database with more than 10 million images comprising $400+$ unique scene categories. MiniPlace is a scaled-down version of Places2, with 100,000 images of 100 categories with a resolution of $128\times128$.

The IMDB-WIKI dataset is collected and published by Rothe et al.\cite{Rothe-IJCV-2016}, which is by far the largest publicly available dataset of face images with gender and age labels for training. The dataset consists of public celebrity facial images with age and gender labels inferred from associated timestamps and names, and totals at 523,051 images. As some images contain multiple faces or suffer from low quality issues, the entire dataset is internally cleaned and preprocessed, as well as merged with internal employee profile image data to enhance its size ($\sim1$ million images) and quality. In this study, only age labels are used.

As one of the most common issues encountered in business applications is the lack of data, where a typical data set would be on the order of 10k to 100k records, only a subset (100k) of both datasets are used throughout the experiments to investigate the effects of data on transfer learning in a similar data volume regime.
\subsection{Neural Network Models}
Different Neural Network architectures are tested in this study. Google's InceptionV3 model\cite{ref8} pre-trained on ImageNet (2012) is tested on the Dogs/Cats dataset as an exploratory analysis on impact of data size. The VGG-Face model\cite{ref6} pre-trained on public internet data is applied to the IMDB-WIKI dataset as an example of applications with highly similar target and source domains. In addition, the VGG-Face model is also applied to the MiniPlace data to investigate transfer learning applications with highly different target and source domains. The VGG16 model\cite{ref9} pre-trained using ImageNet data is applied to the MiniPlace data as an example use case with moderately different target and source domains.

\subsection{Training Process}
A series of transfer learning experiments with different target/source domains and varied training data sizes are conducted to shed lights on the effect of data size on transfer learning performances. The experiments can be  categorized into three groups depending on how different the target domain is form the source domain.

Throughout this paper, two transfer learning strategies are utilized and studied, as defined below:
 
\textit{Feature extraction}: Using a ConvNet pre-trained on source domain data, we remove the last fully connected layer and output layer.  Those layers are then replaced with a new fully connected layer and target domain specific output layer.  The remaining layers are used as a fixed feature extractor, while only the replaced layers are trained.
 
\textit{Fine-tuning}: After accomplishing the training steps specified in feature extraction, we then allow the pretrained layers of the ConvNet to be updated via back-propagation, fine tuning their weights for the target domain. It is possible to fine-tune all the layers of the ConvNet, or  to keep some of the earlier layers fixed and only fine-tune some higher-level portion of the network.

For highly similar source/target domains, the VGG-Face model is trained on the IMDB-WIKI dataset with increasing data size from 10k to 100k incremented by 10k, to predict the individuals' ages. Fine-tuning and feature extraction are respectively applied to each of the experiments.

For moderately different source/target domains, the VGG16 model is trained on the MiniPlace dataset with increasing data size from 10k to 100k incremented by 10k, to classify the different scenes contained in the data. Fine-tuning and feature extraction are respectively applied to each of the experiments.

For highly different source/target domains, the VGG-Face model is trained on the MiniPlace dataset with increasing data size from 10k to 100k incremented by 10k (random sampling), to classify the different scenes contained in the data. Fine-tuning and feature extraction are respectively applied to each of the experiments.

All training processes utilized the ``ADAM''\cite{ref10}  optimizer with number of epochs set to 10 to ensure consistent comparisons across all models. Model performances are evaluated on a holdout dataset of 10k in size. The `categorical accuracy' metric is used for evaluating classification models, while the `mean absolute error' metric is used to evaluating regression models.

\section{Results and Discussions}
\subsection{Impact of Data Size}
\begin{figure}[h]
	\begin{subfigure}{0.5\linewidth}
		\centering
		\includegraphics[width=1\textwidth]{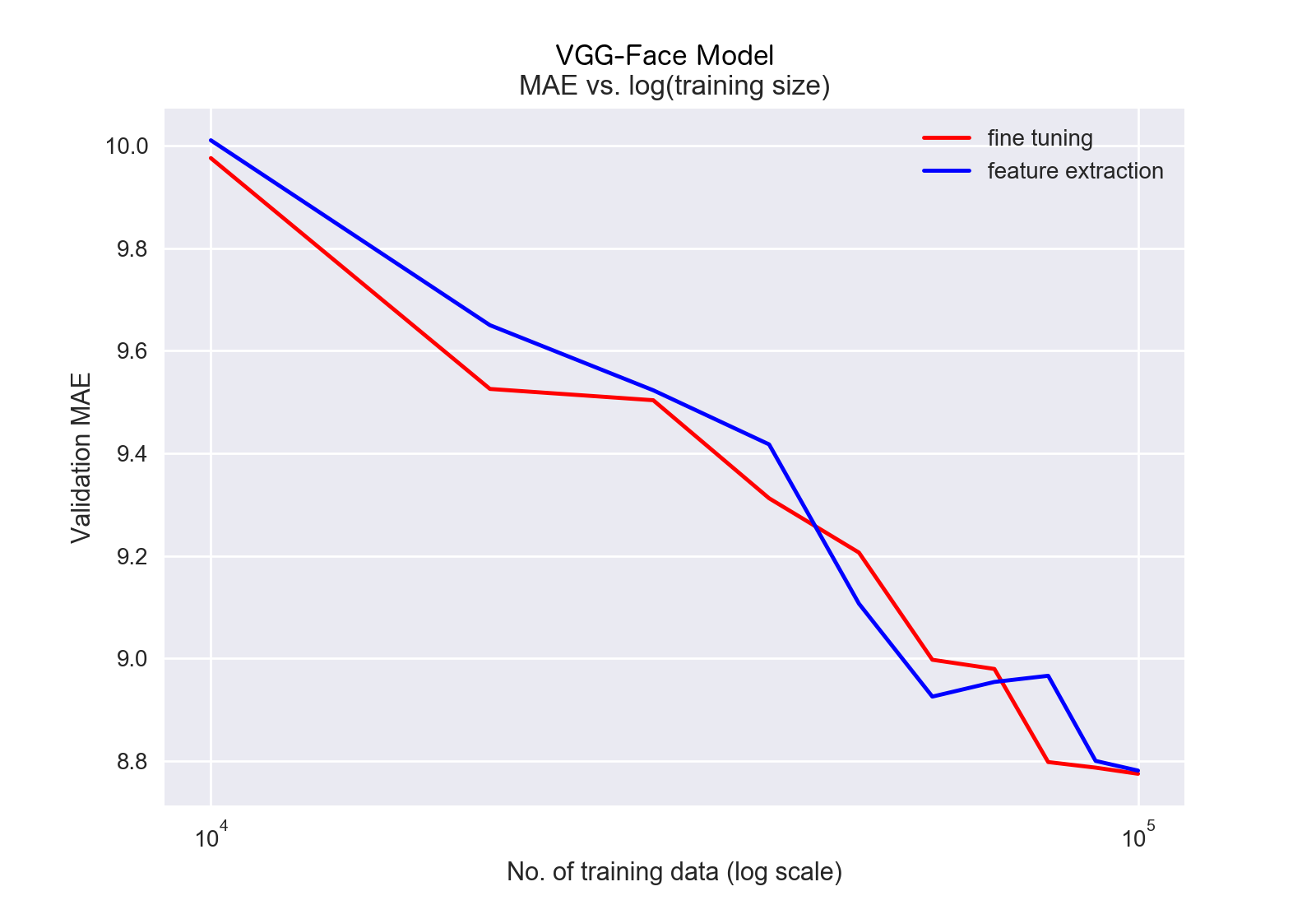}
		\caption{Source: VGG-Face; Target: IMDB}
	\end{subfigure}
	\begin{subfigure}{0.5\linewidth}
		\centering
        		\includegraphics[width=1\textwidth]{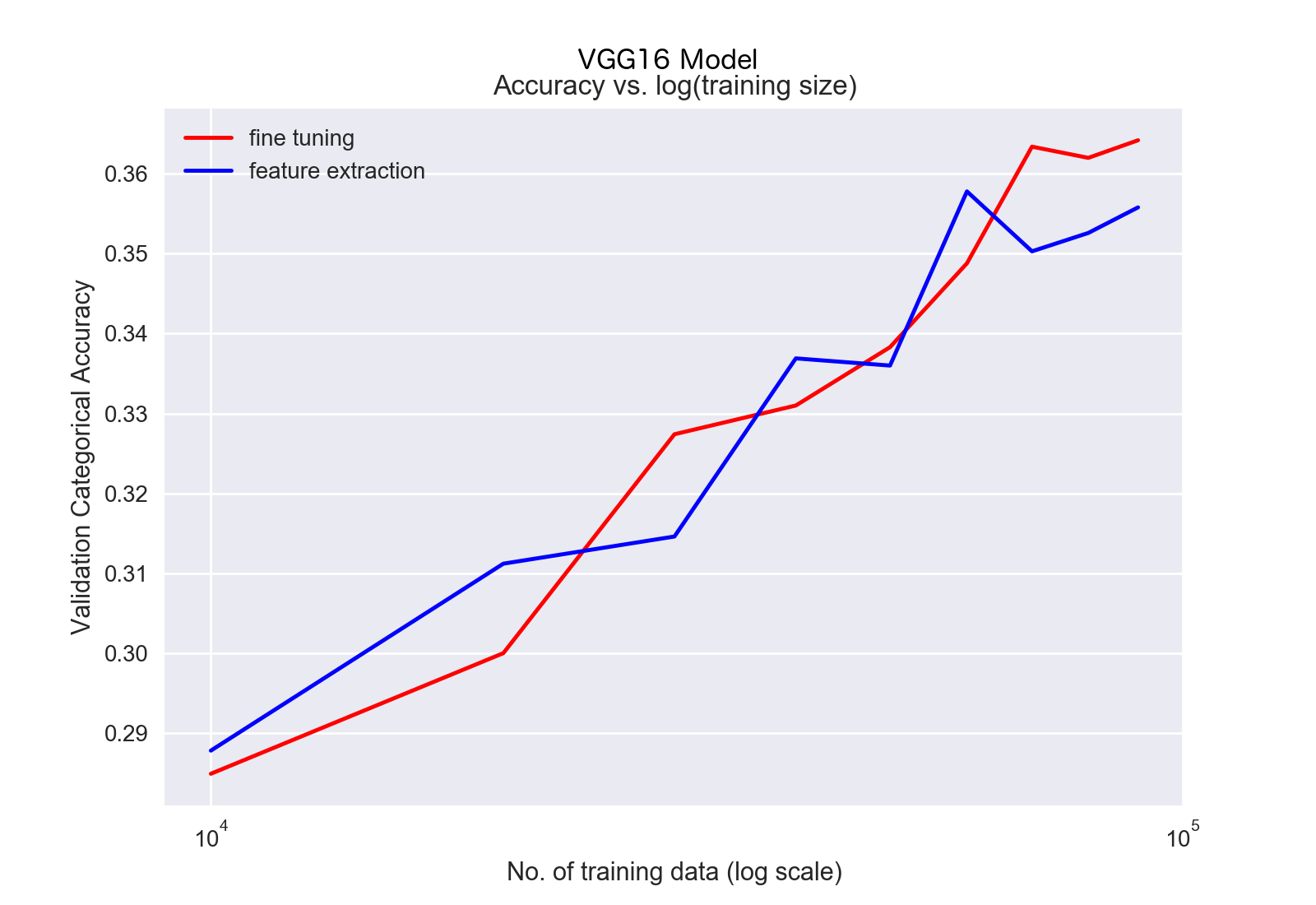}
		\caption{Source: VGG16; Target: MiniPlace}
	\end{subfigure}
	\begin{subfigure}{\linewidth}
		\centering
		\includegraphics[width=0.5\textwidth]{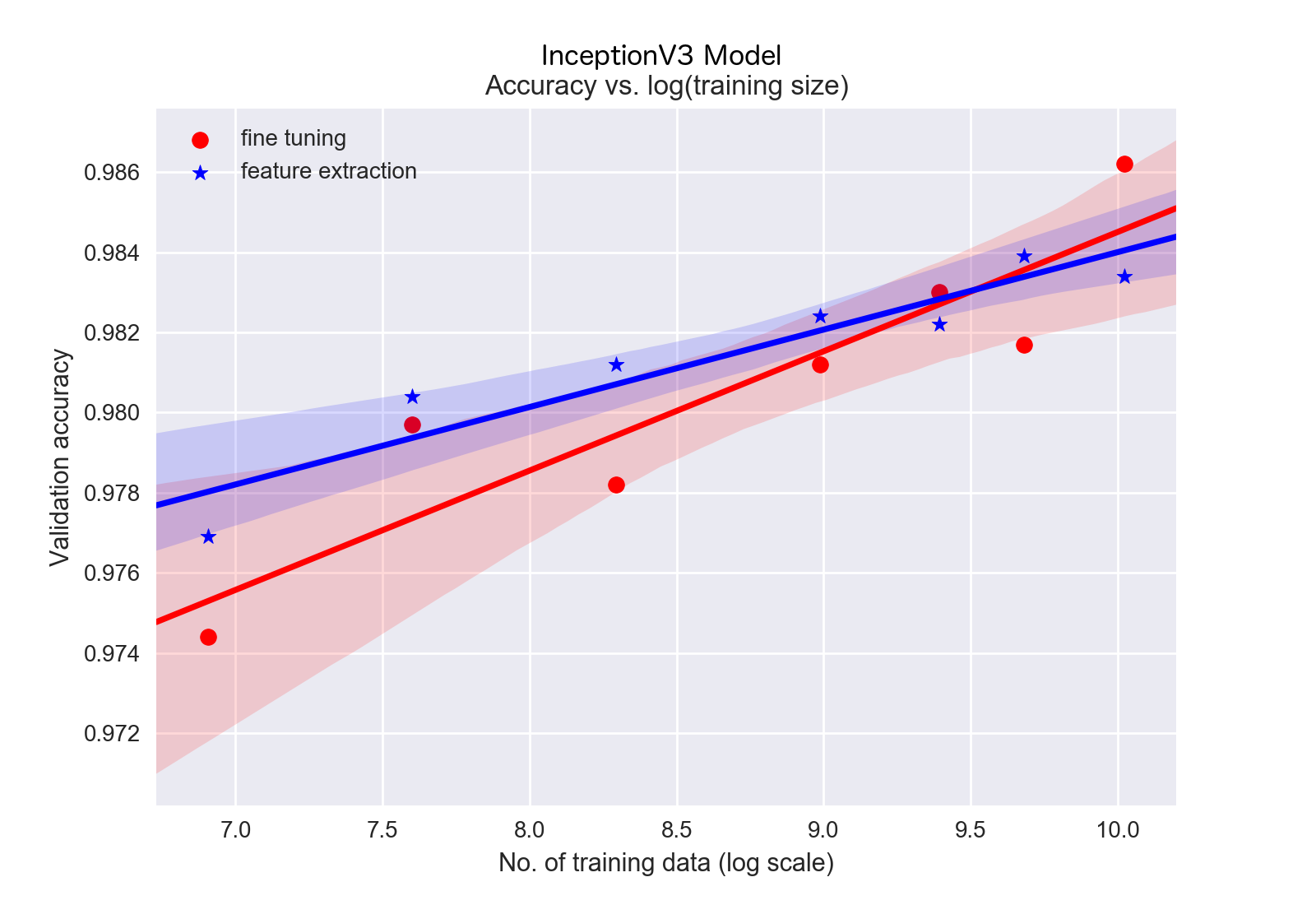}
		\caption{Source: Inception V3; Target: Dogs/Cats}
	\end{subfigure}
	\caption{Model performance is logarithmically correlated to training data size. The VGG-Face model is shown on the top left, while the VGG16 model is shown on the top right. The lower left shows the results from the InceptionV3 model}
	\label{fig:fig1}
\end{figure}

For this experiment, all model parameters are fixed except for the size of the training data, which grows from 10k to 100k with an increment step of 10k. The experiment is conducted on VGG-Face, VGG16, and Inception V3 models with IMDB-WIKI, MiniPlaces, and Dogs/Cats datasets, respectively. Figure~\ref{fig:fig1} shows the improvement of model accuracy as the training data size grows. Not surprisingly, with the same number of epochs and learning rate, the model accuracy continues to increase as training data expands, which demonstrates the effectiveness of data size. In addition, the lift in model performance scales logarithmically with the data size, in consistency with the findings in \cite{ref3}.

\subsection{Highly Similar Target and Source Domains} 
For this experiment, fine-tuning and feature extraction are each applied to training the VGG-Face model with the IMDB-WIKI datasets to predict individuals' ages in the images. The models are trained with 10 epochs with expanding training data from 10k to 100k. Since the original data used in the VGG-Face model and the IMDB-WIKI dataset both consist of facial images, this experiment serves as an investigation into transfer learning applications with highly similar target/source domains.

\begin{figure}[h]
	\centering
	\begin{subfigure}[b]{\columnwidth}
		\includegraphics[width=0.48\textwidth]{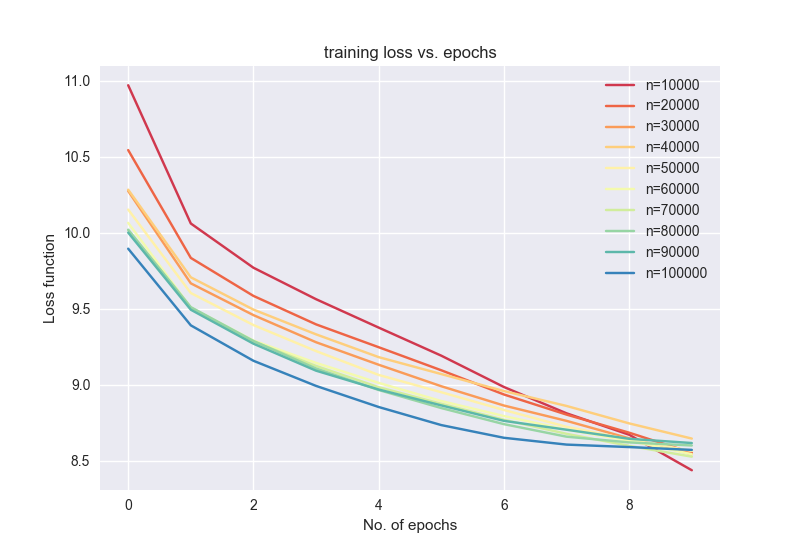}
		\includegraphics[width=0.48\textwidth]{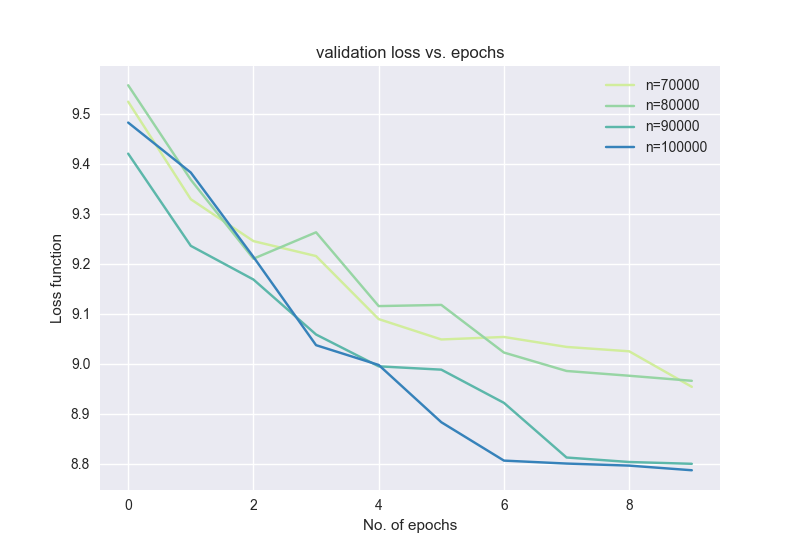}
		\caption{Feature extraction}
	\end{subfigure}
	\begin{subfigure}[b]{\columnwidth}
		\includegraphics[width=0.48\textwidth]{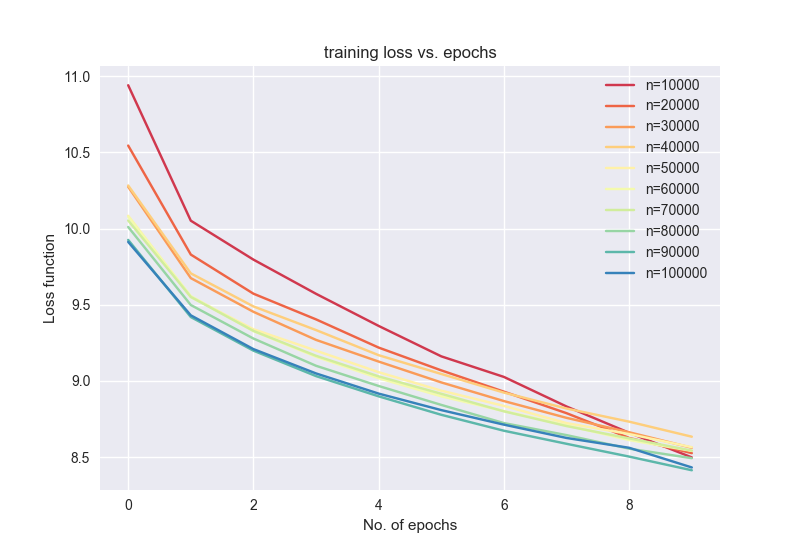}
		\includegraphics[width=0.48\textwidth]{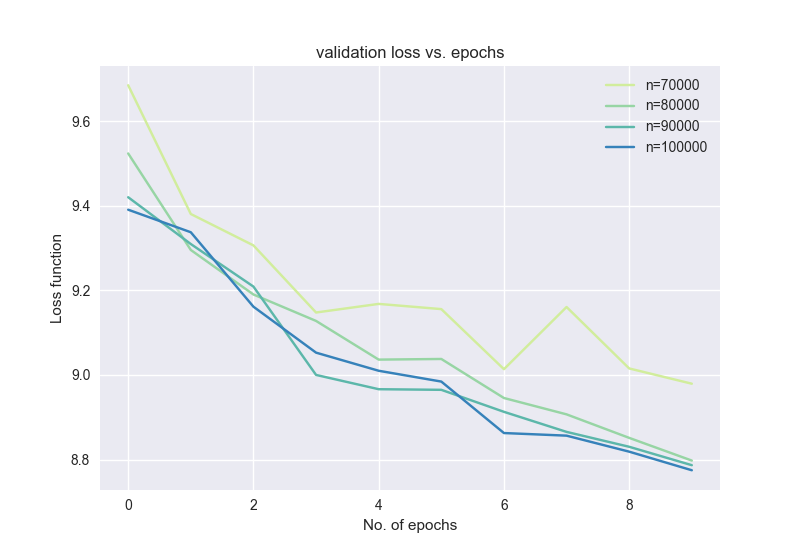}
		\caption{Fine-tuning}
	\end{subfigure}
	\caption{Source: VGG-Face; Target: IMDB-WIKI. Comparison of feature extraction and fine-tuning in reduction of loss function values. The decreasing rates of both training and validation loss diminish as training data grows under feature extraction, while the losses keep declining in fine-tuning.}
	\label{fig:fig2}
\end{figure}

Figure~\ref{fig:fig2} shows the decay of loss function over the training epochs for each experiment, where results from feature extraction are illustrated in the left two subfigures, and fine-tuning is shown to the right.
It is evident that both the training and validation loss values using feature extraction tend to stabilize and plateau by the end of 10 epochs, as more training data become available. Seemingly, the transition point is roughly when training data contains 80k images. Contrarily, fine-tuning leads to an ever-decreasing train and validation loss indicating the model performance can be further improved if more data and training time are provided. Such observations indicate that for transfer learning problems with highly similar target and source domains, it would be beneficial to perform fine-tuning rather than feature extraction when the training data becomes large enough. For this particular combination of VGG-Face model and IMDB-WIKI datasets, the transition training size is around 80k. However, it is possible a corresponding training size exists for every transfer learning setting, above which fine-tuning gives better model performance while feature extraction can still produce similar results when the training data are below the transition point. 

\subsection{Moderately Different Target and Source Domains}
For this experiment, fine-tuning and feature extraction are each applied to training the VGG16 model with the MiniPlace data, to classify image scenes in the data. The models are trained with 10 epochs with expanding training data from 10k to 90k. ImageNet is the source data VGG16 is trained on, which contains  thousands of image categories including some similar scenes in the MiniPlace data, while other categories differ significantly from the MiniPlace images. Therefore, this experiment serves as an investigation into transfer learning applications with target/source domains sharing limited similarities, but still being largely different.

\begin{figure}[H]
	\centering
	\begin{subfigure}[b]{0.48\columnwidth}
		\includegraphics[width=\textwidth]{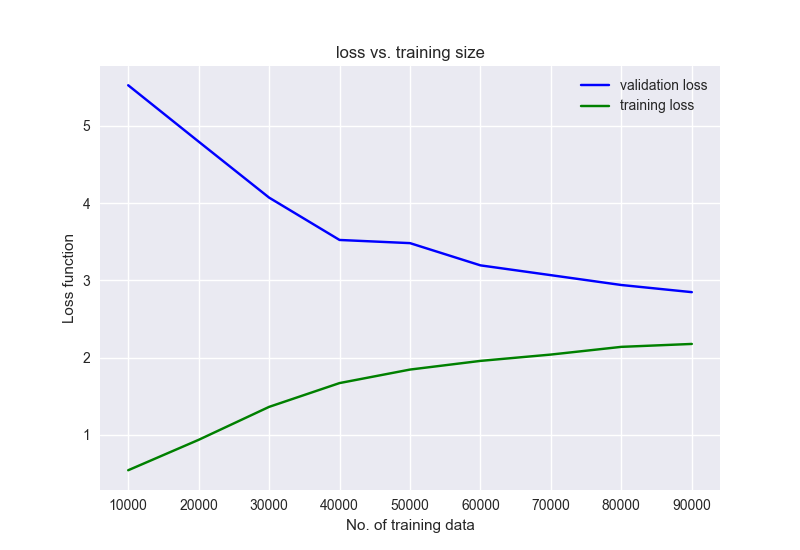}
	\end{subfigure}
	\begin{subfigure}[b]{0.48\columnwidth}
		\includegraphics[width=\textwidth]{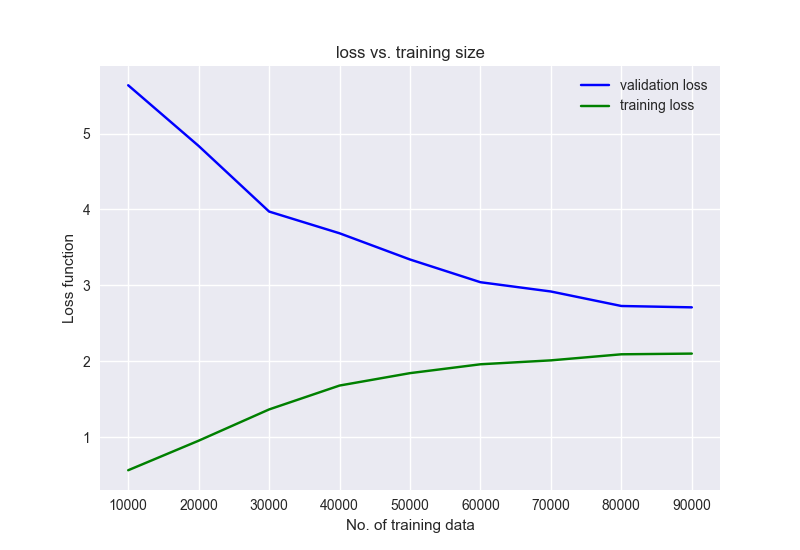}
	\end{subfigure}
	\caption{Source: VGG16; Target: MiniPlace. Comparison of feature extraction and fine-tuning in reduction of loss function values. The first row panel shows the results of feature extraction, while the second row panel shows the results of fine-tuning. Both training and validation loss are slightly lower in fine-tuning than feature extraction by the end of training.}
	\label{fig:fig3}
\end{figure}

The experimental results are shown in Figure~\ref{fig:fig3} with feature extraction plotted on the first row and fine-tuning on the second row. These plots compare the evolution of both training validation with respect to training epochs under both feature extraction and fine-tuning. It is apparent that there is a sizable gap between training and validation loss, which are also large in absolute values in both plots. This implies the model suffers from higher bias compared to the experiment using IMDB-WIKI datasets, meaning that training becomes more difficult as target and source data domains diverge. Furthermore, the final loss values in both training and validation are lower in the case of fine-tuning than feature extraction, although the difference is not so obvious. 

It is observed that fine-tuning renders similar training and validation losses compared to feature extraction, when the training size is small; when training size gets more than 60,000, fine-tuning starts to out-perform feature extraction with lower validation losses generated. Additionally, overfitting is present in both experiments, where training loss increases as the size of training data increases monotonically, which can be explained by the small size of training data, noting we deliberately compared the performance of fine-tuning and feature extraction across different sizes from small to large in experiments.

\subsection{Highly Different Target and Source Domains}
For this experiment, fine-tuning and feature extraction are each applied to training the VGG-Face model with the MiniPlace dataset, to classify image scenes in the data. The models are trained with 10 epochs with expanding training data from 10k to 90k. As aforementioned, the source data used in VGG-Face model primarily contains facial information, distinctively different than the MiniPlace data which contains general scenes such as airport and county yard, this experiment is hence an investigation in to transfer learning applications with target/source domains dramatically different from each other.

\begin{figure}[H]
	\centering
	\begin{subfigure}[b]{\columnwidth}
		\includegraphics[width=0.48\textwidth]{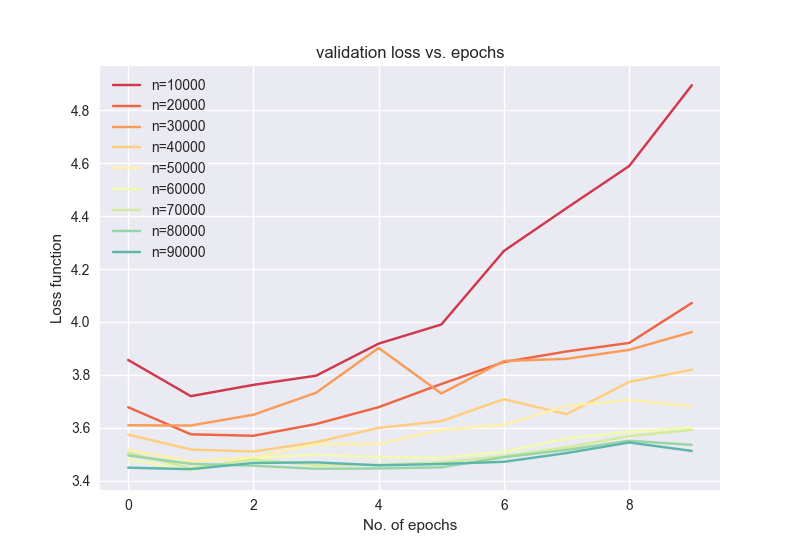}
		\includegraphics[width=0.48\textwidth]{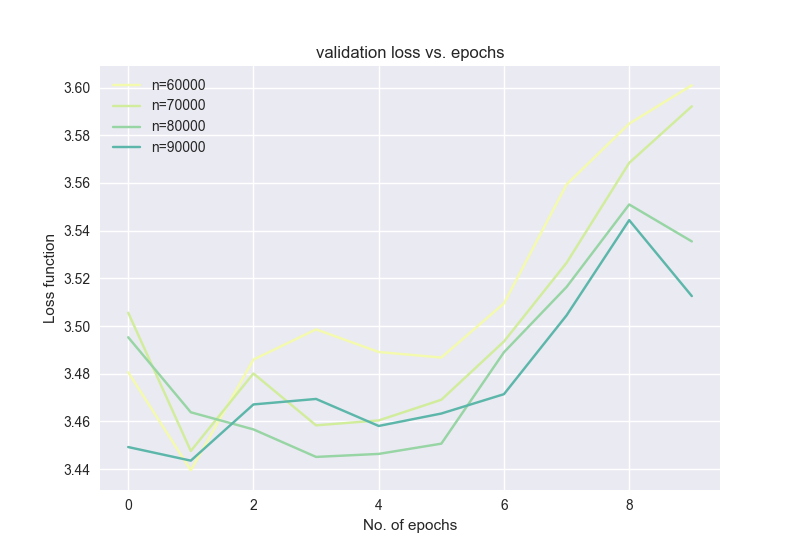}
		\caption{feature extraction}
	\end{subfigure}
	\begin{subfigure}[b]{\columnwidth}
		\includegraphics[width=0.48\textwidth]{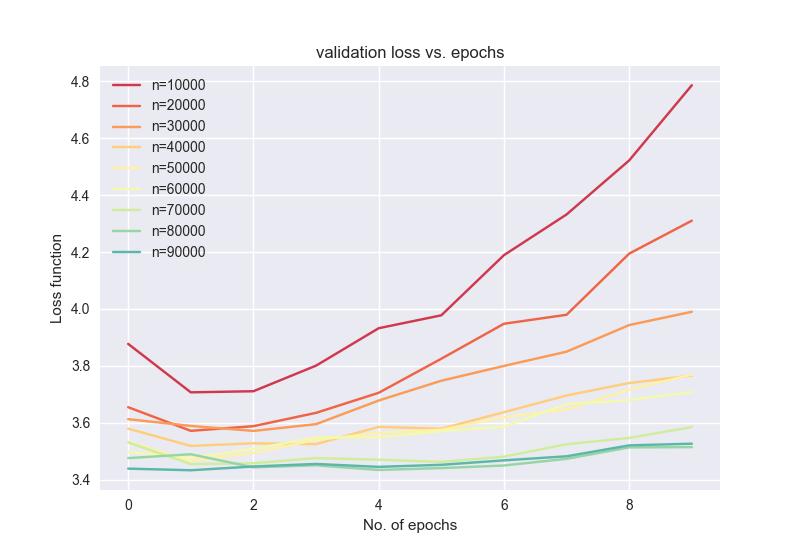}	
		\includegraphics[width=0.48\textwidth]{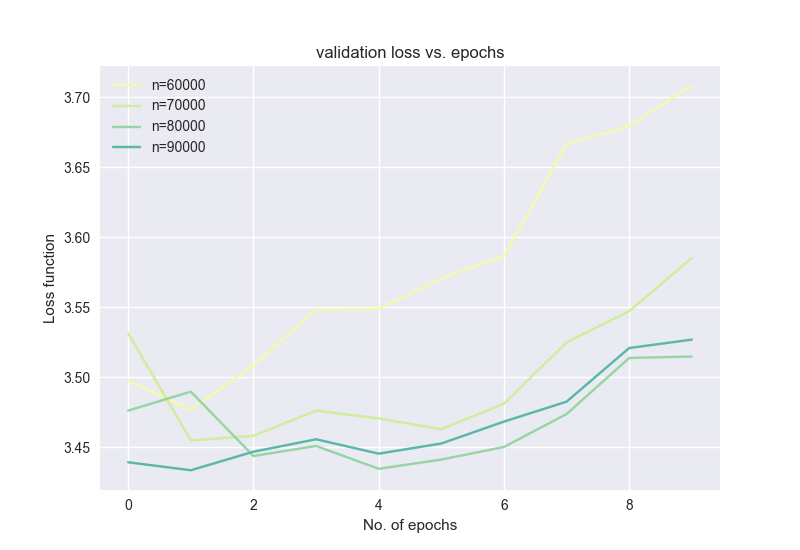}
		\caption{Fine-tuning}
	\end{subfigure}
	\caption{Source: VGG-Face; Target: MiniPlace. Comparison of feature extraction and fine-tuning in reduction of loss function values. A dip in validation loss starting around training epoch 3 is observed when training size is larger than 60k in both sets of experiments.}
	\label{fig:fig4}
\end{figure}

Results from both fine-tuning and feature extraction are summarized in Figure~\ref{fig:fig4}, showing the decay of validation loss function over the training epochs for each experiment. Feature extraction is plotted on the first row, whereas fine-tuning is shown on the second row. Compared to the last experiment, it becomes more difficult to achieve similar accuracies with the same number of epochs in training. As the target/source domains become more different, the level of difficulty in training increases.
 
Although the validation loss increases over training epochs (an implication of overfitting) and no obvious differences in model performance are observed between feature extraction and fine-tuning, there seems to be a critical training size of approximately 60k, where validation loss exhibits a dip starting between training epoch 3 and 4. In addition, as the training size becomes larger, the increase in validation loss slows down and eventually starts decreasing when training size approaches 80k or 90k. Such observations indicate that early stopping may be beneficial in this problem setting when data size is limited. Having larger training data is the key to better model performance and 100k is still not large enough to uncover performance differences between feature extraction and fine-tuning. 

\section{Conclusions}
In this study, we investigate the impact of target data size in a typical business application of transfer learning where training data are very much limited. In addition, we study how similarity between target and source data domains could affect the final model performance by using different transfer learning training approaches, namely feature extraction and fine-tuning. From a series of experiments, we find that in general, a larger target data size will always lead to better training convergence, although the marginal performance decreases as the dataset grows. More precisely, there is an apparent logarithmic relationship between model performance and the target data sizes. The level of similarity between the source and target data domains also plays a vital role in selecting the appropriate training framework. When target and source domains are similar, feature extraction gives equivalent training results as fine-tuning in small training data settings. However, fine-tuning tends to give better results as data size increases, and each specific target/source combination likely has a unique such transition data size. When target and source domains diverge, it's crucial to collect more training data. While both feature extraction and fine-tuning are comparable throughout all experiments, the validation performance will only increase when target training data are sufficiently large.

\section{Discussion}
Data is a precious asset, but unfortunately data is often in short supply and of low quality.  Our paper hopes to begin a quantitative discussion around data volume and transfer learning target domain performance. 
 
Most of the results we discussed in this paper aren?t shocking, but to the best of our knowledge this is the first time these ideas have been quantitatively explored. 
 
More data seems to always produce some gain in model performance, and model performance tends to increase with the log of data volume. There is a cost to gathering, curating, and processing that data that can be considered against this performance gain.
 
The similarity between source and target domain plays a big part in the amount of data required for a transfer learning application. As the target and source domains become less similar we see that the amount of data required for some performance target increases and we also observed that fine tuning on the target domain becomes more important.  Imagining the layers of the transferred model as specialized feature extractors, it seems reasonable that differing source and target domains would cause these feature extractors to be less suitable for the target domain task, which likely leads to this consideration for fine tuning.
 
Our hope is that by beginning this quantitative investigation on the effect of data volume and domain similarity in transfer learning we might inspire others to explore the significance of data in developing more accurate statistical models.

%
%

\end{document}